\documentclass{article}
\usepackage[accepted]{icml2022}

\usepackage[utf8]{inputenc}
\usepackage[T1]{fontenc}
\usepackage{hyperref}
\usepackage{url}
\usepackage{booktabs}
\usepackage{amsfonts}
\usepackage{nicefrac}
\usepackage{microtype}
\usepackage{xcolor}

\usepackage{graphicx}
\usepackage{subcaption}
\usepackage{amsmath}
\usepackage{amssymb}
\usepackage{mathtools}
\usepackage{amsthm}
\usepackage{bm}
\usepackage{algorithmic}
\usepackage{algorithm}
\usepackage{wrapfig}
\usepackage{multicol}
\usepackage{listings}
\usepackage{pythonhighlight}
\usepackage[capitalize,noabbrev]{cleveref}
\usepackage[textsize=tiny]{todonotes}

\theoremstyle{plain}

\theoremstyle{definition}

\theoremstyle{remark}

\newcommand{\nan}{\varnothing}
\newcommand{\ts}{\mathbf{x}}
\newcommand{\newts}{\mathbf{y}}
\newcommand{\spline}{\mathcal{S}_\ts}
\newcommand{\newspline}{\mathcal{S}_\newts}
\newcommand{\diff}{\mathrm{d}}
\newcommand{\order}{k}

\def\Figref#1{Figure~\ref{#1}}

\def\eqref#1{Eq.\ \ref{#1}}

\def\1{\bm{1}}

\def\va{{\bm{a}}}
\def\vb{{\bm{b}}}
\def\vc{{\bm{c}}}

\def\vx{{\bm{x}}}

\def\mW{{\bm{W}}}

\DeclareMathAlphabet{\mathsfit}{\encodingdefault}{\sfdefault}{m}{sl}
\SetMathAlphabet{\mathsfit}{bold}{\encodingdefault}{\sfdefault}{bx}{n}

\newcommand{\R}{\mathbb{R}}

\hypersetup{
  colorlinks,
  linkcolor={red!50!black},
  citecolor={blue!50!black},
  urlcolor={blue!80!black}
}

\newcommand{\theTitle}{Irregularly-Sampled Time Series Modeling with Spline Networks}
\icmltitlerunning{\theTitle}

\begin{document}

\onecolumn
\icmltitle{\theTitle}
\icmlsetsymbol{equal}{*}

\begin{icmlauthorlist}
\icmlauthor{Marin Bilo\v{s}}{}
\icmlauthor{Emanuel Ramneantu}{}
\icmlauthor{Stephan G\"{u}nnemann}{}\\
Technical University of Munich, Germany
\end{icmlauthorlist}

\icmlcorrespondingauthor{Marin Bilo\v{s}}{marin.bilos@tum.de}

\icmlkeywords{} 

\vskip 0.3in

\printAffiliationsAndNotice{}

\begin{abstract}
  Observations made in continuous time are often irregular and contain the missing values across different channels. One approach to handle the missing data is imputing it using splines, by fitting the piecewise polynomials to the observed values. We propose using the splines as an input to a neural network, in particular, applying the transformations on the interpolating function directly, instead of sampling the points on a grid. To do that, we design the layers that can operate on splines and which are analogous to their discrete counterparts. This allows us to represent the irregular sequence compactly and use this representation in the downstream tasks such as classification and forecasting. Our model offers competitive performance compared to the existing methods both in terms of the accuracy and computation efficiency.
\end{abstract}

\section{Introduction}\label{sec:introduction}

\begin{wrapfigure}{r}{0.41\textwidth}
    \centering
    \vspace{-0.9cm}
    \input{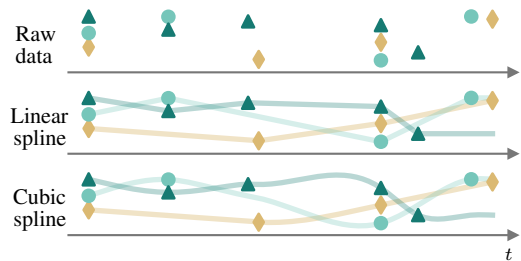}
    \vspace{-0.7cm}
    \caption{Interpolating irregular data using splines.}
    \label{fig:spline_example}
    \vspace{-0.7cm}
\end{wrapfigure}

Irregularly-sampled data with missing values can be found in many domains. For example, medical records are a collection of multivariate attributes in continuous time, where only a small subset of possible measurements are taken at a certain time point. Both the missingness and the irregularity can be addressed by using some sort of interpolation on the raw data.
One popular option is using splines, i.e., piecewise polynomials that fit the input points \cite{moritz2017imputets}. The missing points can be imputed by simply evaluating the polynomial function at the target time points. \Figref{fig:spline_example} demonstrates this: the original data is irregular and contains the missing values but after fitting the spline we can fill in the gaps. Additionally, we can evaluate the spline at any time point between the observed values to infer about the areas where we have not seen any data in the first place.\looseness=-1

The existing methods such as RNNs \cite{che2018recurrent}, and neural ordinary and controlled differential equations \citep{chen2018neural,kidger2020neural} use the imputed points but still need to take care of the irregular sampling. In their case, the latent state is evolved continuously over time. This approach is usually slower and the models are more likely to struggle with long-term dependencies \citep{lechner2020learning}, just like their discrete counterparts.

We propose an alternative where the imputed points are never materialized; rather, we work on the spline representation of the time series directly since it uniquely represents the data. That is, we use splines as an input to a neural network which in turn outputs a modified spline. We develop the layers that act as a continuous version of discrete layers, such as $1 \times 1$ convolution and pooling operations.
The final output of our model is the latent representation of the underlying time series, in the functional form. We can use this to obtain the unique fixed-sized latent vector that represents the whole time series, by simply querying the spline interpolation at the desired time points. The latent vector can be further used in the downstream tasks.

\section{Method}\label{sec:model}

We represent the time series as a sequence of observation-time pairs $\ts = ((\vx_1, t_1), \dots, (\vx_n, t_n))$ where $t_i \in [0, T]$ is the time at which $\vx_i \in (\R \cup \{\mathrm{\nan}\})^d$ is observed, with the symbol $\nan$ denoting the missing values.
Given $\ts$, we fit a spline to the observed data points to obtain the interpolation function $\spline : [0, T] \rightarrow \R^d$. Spline $\spline$ consists of a sequence of $n - 1$ polynomials $P_i(t)$, each polynomial is defined on the interval between two consecutive observations, also called the knots. More precisely, we define a $k$th order polynomial between the observations $\vx_i$ and $\vx_{i+1}$ as $\smash{P_i : [t_i, t_{i+1}] \rightarrow \R^d}$, $\smash{P_i(t) = \sum_{i=0}^k \va_i t^i}$, where $\va_i \in \R^d$ are the coefficients.

A simple example is a linear spline that connects the knots $(\vx_i, t_i)$ with straight lines. The coefficients of the linear function $P_i(t) = \va_1 t + \va_0$ can be easily calculated from data as $\va_1 = \frac{\vx_{i + 1} - \vx_i}{t_{i + 1} - t_i}$ and $\va_0 = \vx_i - \va_i t_i$. Besides linear, we can have piecewise constant (rectilinear) function that copies the previous value until the next observation, or higher order polynomials, cubic being the most popular one \cite{de1978practical}. For cubic splines we require that the derivative in the knots matches from both sides, i.e., that they are smooth. We get the coefficients by solving the system of equations that imposes this condition.

To evaluate the spline at an arbitrary time point $t$, we first need to find the interval where $t_i \leq t \leq t_{i+1}$ holds, then compute $P_i(t)$. Consequently, evaluating $\spline(t)$ at the observed time points $(t_1, \dots, t_n)$ performs the imputation of the missing values. At this point the previous methods \citep[e.g.,][]{kidger2020neural} would sample the points from the spline and use them in their models. We, instead, process the spline directly to obtain the useful \textit{continuous-in-time} features which, in turn, can be used to obtain, e.g., the fixed sized latent representation of the time series.

\subsection{Spline operations}

\begin{figure*}
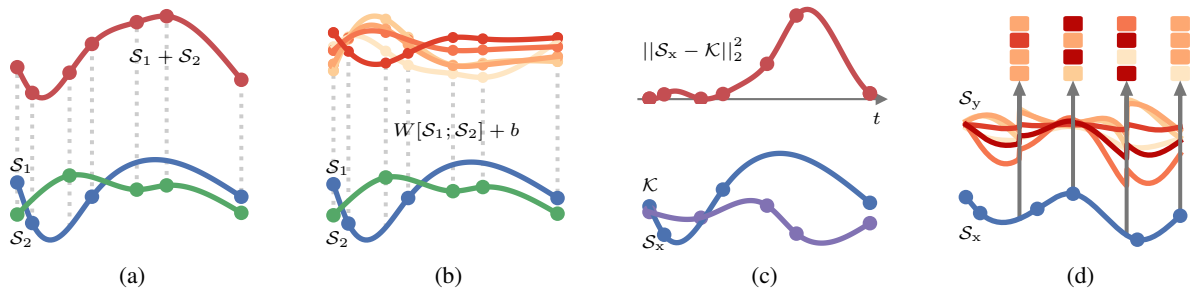

    \centering
    \begin{subfigure}[b]{0.24\textwidth}
        \centering
        \input{figures/spline_sum.pgf}
        \vspace{-0.3cm}
        \caption{}
        \label{fig:spline_sum}
    \end{subfigure}
    \begin{subfigure}[b]{0.24\textwidth}
        \centering
        \input{figures/spline_linear_layer.pgf}
        \vspace{-0.3cm}
        \caption{}
        \label{fig:spline_linear_layer}
    \end{subfigure}
    \begin{subfigure}[b]{0.24\textwidth}
        \centering
        \input{figures/spline_distance.pgf}
        \vspace{-0.3cm}
        \caption{}
        \label{fig:spline_distance}
    \end{subfigure}
    \begin{subfigure}[b]{0.24\textwidth}
        \centering
        \input{figures/splinenet.pgf}
        \vspace{-0.3cm}
        \caption{}
        \label{fig:splinenet}
    \end{subfigure}
    \caption{(\subref{fig:spline_sum}) Summing up two splines is equal to summing the polynomial coefficients on the shared intervals. (\subref{fig:spline_linear_layer}) An affine layer mixes the channels and projects the spline to a latent space. (\subref{fig:spline_distance}) The distance between two splines is again a spline. We can use a learnable spline $\mathcal{K}$ to obtain the interpretable features. (\subref{fig:splinenet}) The SplineNet model has the option to split the input spline into smaller segments allowing it to extract the features locally. Each of the depicted four segments produces a single latent vector that we can aggregate, e.g., using a recurrent neural network.}
    \vspace*{-0.8cm}
\end{figure*}

The operations on splines often reduce to operations on polynomials. For example, adding two splines that share the same set of time points $(t_1, \dots, t_n)$ results in adding the coefficients on the corresponding intervals. If the splines do not share the time points, we have to add the missing knots and define polynomials on newly generated intervals, as is shown in \Figref{fig:spline_sum}.

Another useful operation is the multiplication between two splines. This is again done piecewise. Multiplying two $\order$th degree polynomials requires $O(\order^2)$ operations but can be reduced to $O(\order \log \order)$ \citep[see, e.g.,][ch.\ 30]{cormen2009introduction}. The key insight is that the polynomial multiplication can be seen as a convolution.  Thus, we can use the convolution theorem, i.e., apply the fast Fourier transform to the coefficients and perform the convolution as a pointwise product. This comes at a reduced computation cost compared to the naive approach. We get the resulting polynomial coefficients via inverse Fourier transform.

Now we tackle matching the knots of two arbitrary splines such that the result is two splines which share the knot positions exactly.
If the spline has $n$ knots and we want to add a new knot at time $t$, we can simply find the interval $[t_i, t_{i+1}]$ on which $t$ falls and construct two new intervals $[t_i, t]$ and $[t, t_{i+1}]$. The first interval keeps the unchanged polynomial coefficients while the second one needs to shift the polynomial by $t - t_i$.
We shift the polynomials using the so-called Taylor shift algorithm \cite{von1997fast} which has the complexity $O(k \log k)$ for a $k$th degree polynomial. In general, when we have an operation between two splines (e.g., a sum) with $n$ and $m$ knots, the worst case scenario gives us a spline with $n + m$ knots.

\subsection{SplineNet}

Given a spline $\spline$, we wish to transform the individual points using an \textbf{affine layer} $\newspline(t) = \mW \spline(t) + \vb$. The output function $\newspline$ will again be a spline with the same order as $\spline$. This is easy to show since the affine function is multiplying the polynomials by a constant, adding them together, and finally, adding the constant to the result. Therefore, we can implement an affine layer by only handling the polynomial coefficients such that we never have to materialize the actual data points.
Using such a layer allows for \textit{mixing} between the time series channels (see \Figref{fig:spline_linear_layer}). The resulting spline is defined with knots $\newts$ that lie on the same time points as in $\ts$, however, they have different values and dimension. This layer, in essence, projects the original interpolated time series to a continuous latent interpolation. It is a continuous version of the $1 \times 1$ convolutional neural network.

An affine layer as defined above ignores the temporal dependencies since it processes points independently of their neighbors. Thus, we would like to have a way of aggregating the information over time. One approach is to perform the \textbf{integration} --- a continuous equivalent of sum-pooling. That is, we wish to obtain $\mathcal{I}_\newts(t) = \smash{\int_0^t \newspline(\tau) \diff \tau}$. Since $\newspline$ is a piecewise function, we can integrate over the individual sub-functions and add up the results. That means we need to calculate the integral of an arbitrary polynomial function $P(t) = \sum_{i=0}^\order \vc_i t^i$ which is simply $\sum_{i=0}^{\order + 1} \frac{\vc_{i - 1}}{i} t^i$ where $\vc_{-1}$ is introduced as an the offset that is calculated based on the value of the previous sub-functions. The resulting curve $\mathcal{I}_\newts(t)$ is again a spline with an order $\order + 1$.\looseness=-1

Since integrating can lead to large values, we introduce a version of the \textbf{batch normalization layer} \cite{ioffe2015batch} that normalizes the area under the curve to have the mean equal to $1$. The area statistics can be easily calculated by evaluating the integral spline at the last time point.
Affine and integration layers act across the different channels and along the time dimension. For certain downstream tasks we want to represent the time series and its latent spline with a finite dimensional vector. One way to do this is to query the spline at the last observed time point $t_n$. In this case, the proposed layers cannot distinguish certain splines, for example, integration aggregates the data without caring about the order in which it arrived.

To alleviate this, we add another type of layer where we multiply the input spline with a \textbf{learnable spline kernel}. The kernel is defined with learnable knot locations and polynomial coefficients. The result is again a spline. This corresponds to a convolution in the frequency domain. In practice, we use multiple kernels in parallel to extract more features.
An alternative to multiplying the spline $\newspline$ with a spline kernel $\mathcal{K}$ is to use the kernel to learn the shape of the spline. That is, the output is now defined as the distance between the two splines: $|| \newspline - \mathcal{K} ||_2^2$ (see \Figref{fig:spline_distance}). Integrating over this path gives us the distance between the two curves. The model essentially defines \textbf{learnable shapes} $\mathcal{K}$ which it tries to match with the input data shapes, giving us an interpretable layer, similar to the idea of shapelets \cite{ye2009shapelet}.

Equipped with these layers, we can extract useful information from the original time series. However, if the time series is very long, the model has to learn the whole shape, but in practice, it might be better to extract the features locally.
We can split the time series $\ts$ into $l$ smaller equally-spaced segments, apply the same layers as above and obtain $l$ latent vectors. These vectors can then be passed to an RNN or some other established model for \textit{regular} time series to get the unique latent representation of the whole input. Again, in practice, we do not actually need to split the time series; instead we query the final latent spline at $l$ time points. The positions of these points can also be learnable.

\subsection{Related work}\label{sec:related}

The irregular data can be found in medicine \citep{mimic4}, astronomy \citep{scargle1982studies}, climatology \citep{schulz1997spectrum}, and signal processing. Most of the approaches fill-in the missing values from the observed data either explicitly or implicitly, e.g., in non-uniform Fourier transform \citep{dutt1993fast,lomb1976least}, decision trees \citep{cutler2012random}, linear regression \citep{morvan2020neumiss}, and neural networks \cite{khosravi2019tractable,smieja2018processing}.
Existing methods that perform interpolation include gaussian processes \cite{rasmussen2003gaussian}, polynomial fitting, neural networks \cite{shukla2019interpolation}, or using piecewise functions. In our work we use splines since they are closed under our proposed model, i.e., the output of the neural network is again a spline.

Some recent approaches extend the RNNs to operate in continuous time \cite{che2018recurrent,neil2016phased,mei2017neural}. \citet{che2018recurrent} introduce an extension of a GRU cell \cite{cho2014gru}, that decays the hidden state between the observations.
With the introduction of neural ODEs \cite{chen2018neural}, several approaches proposed the evolution of the hidden state with a learnable ODE \citep{de2019gru,rubanova2019latent,jia2019jump}.
Since solving an ODE can be expensive, \cite{bilos2021flow} introduce an alternative model that learns the solution curve directly. \citet{kidger2020neural} propose using neural controlled differential equations that use the continuous input to drive the latent representation using a reparameterized ODE. They use splines to construct the continuous path, which is evaluated multiple times during one solver call. \citet{shukla2021transformer} extend transformers \citep{vaswani2017attention} by using the time attention and introduce the reference points on top of which RNN is used to output the final sequence representation. This is similar to our approach of dividing the time series into segments.

Shapelets \cite{ye2009shapelet,grabocka2014learning,kidger2020generalised} are interpretable feature extractors that aim to learn the shape of the input curve by sliding along the time series and computing the distance between the input points and the shapelet, which are used as an interpretable input to a downstream model. In our case, the shapes are learned splines that are fixed in position and potentially repeated across segments. We can also learn the shapes in the latent space which limits the interpretability but increases expressiveness. Since we avoid the expensive shapelet search algorithm, our method does not suffer from performance issues \cite{rakthanmanon2013fast}.

\section{Experiments}\label{sec:experiments}

In this section we demonstrate that our method can be used as an alternative to the established irregularly-sampled time series models through a set of different experiments such as classification, forecasting and probabilistic modeling. All the experiment were done on a single GPU with 12GB of memory.

\textbf{Classification.}
We consider three datasets for our evaluation and process them similar to \cite{kidger2020neural}.
\underline{\smash{Character Trajectrories}} \cite{bagnall2018uea} represents handwritten characters, each sequence has length 182 and 3 channels. To add irregularity, we create two augmented datasets with 30\% and 70\% of missing values, respectively.
\underline{\smash{Physionet}} \citep{physionet} is a medical dataset from hospitalized patients with the goal of predicting sepsis. It consists of 40,336 sequences with 34 time series features. There is a class imbalance, with 7\% of the patients suffering sepsis and it contains many missing values (90\% overall).
\underline{\smash{Speech Commands}}
dataset contains records with 35 classes corresponding to different spoken words, with sequences of length 161, each with 20 channels. All the datasets are normalized and we randomly split them into train, validation and test sets ($60\%-20\%-20\%$).

\textit{Models:}
In order to assess the performance of our model we compared it to the GRU-D \cite{che2018recurrent}, two architectures based on differential equations, GRU-ODE \cite{rubanova2019latent} and Neural CDE \cite{kidger2020neural}, GRU Flow \citep{bilos2021flow}, and the transformer \cite{shukla2021transformer}. \underline{\smash{SplineNet}} model has affine, kernel convolution, integration, and ReLU layers as the building blocks. After applying these layers, we gather a single vector from each segment. This forms a fixed sized sequence which can then be fed to an RNN.

\textit{Results:} In ablation studies, we notice that using a single block composing of an affine and integration layer followed by a convolution is enough to produce good results. Our experiments demonstrate that introducing integration makes by far the biggest difference, while the sizes of segments and shapes indicate that more is better in this case.
Table~\ref{tab:classification_result} shows the results for all the baseline models and our model. Clearly, our model outperforms the others across all datasets. As for computation efficiency, our model is slower than transformers and much faster than ODE based models.

\begin{table*}[t]
    \centering
    \begin{tabular}{lcccc}
    {} & Character (miss.\ 30\%) & Character (miss.\ 70\%) &            Physionet &      Speech Commands \\
    \midrule
    GRU                 &     0.9535 $\pm$ 0.012 &     0.9366 $\pm$ 0.025 &  0.9340 $\pm$ 0.006 &  0.8624 $\pm$ 0.001 \\
    GRU-D               &     0.9535 $\pm$ 0.023 &     0.9378 $\pm$ 0.016 &  0.9359 $\pm$ 0.006 &  0.8577 $\pm$ 0.012 \\
    GRU-ODE / Flow      &     0.9634 $\pm$ 0.015 &     0.9476 $\pm$ 0.001 &  0.9353 $\pm$ 0.004 &  0.8426 $\pm$ 0.024 \\
    Neural CDE          &     0.9407 $\pm$ 0.014 &     0.9546 $\pm$ 0.008 &  0.9352 $\pm$ 0.003 &  0.8608 $\pm$ 0.012 \\
    Transformer         &     0.9785 $\pm$ 0.003 &     0.9715 $\pm$ 0.006 &  0.9181 $\pm$ 0.010 &  0.8325 $\pm$ 0.001 \\
    SplineNet (Ours)    &\textbf{0.9849 $\pm$ 0.004}&\textbf{0.9785 $\pm$ 0.003}&\textbf{0.9400 $\pm$ 0.001}&\textbf{0.8693 $\pm$ 0.001}\\
\end{tabular}

    \vspace*{-0.3cm}
    \caption{Test accuracy mean and standard deviation, computed over 4 runs. For Physionet, we report AUC to account for data imbalance. Higher is better, the best performing model is in bold.}
    \label{tab:classification_result}
    \vspace*{-0.6cm}
\end{table*}

\textbf{Probabilistic modeling.} We define a latent variable model and set the experiment similar to \citep{chen2018neural, rubanova2019latent}. The comparison is between our encoder-decoder architecture based on SplineNet architecture and those using the neural ODEs. We report the reconstruction loss on the Character dataset with 70\% missing values: GRU-ODE achieves 0.3924 $\pm$ 0.004, while our model outperforms it with 0.3088 $\pm$ 0.014.

\textbf{Forecasting.} We would like to encode the time series to predict the future values using the same encoder-decoder architectures as before. We use Physionet and Speech Commands dataset with the forecast window of 10 steps. Our method again outperforms the GRU-ODE model, we obtain reconstruction error of 0.0023 and 0.1736, compared to GRU's 0.0025 and 0.1754, on Physionet and Speech, respectively.

\textbf{Interpretability.}
As we already mentioned, the kernel layer with the distance measure can be used as an interpretable layer since it learns the shapes of the input data, similarly to shapelets \citep{ye2009shapelet}. Thus, we compare to \cite{kidger2020generalised} who generalize the shapelets to the irregular data.
We use the Japanese Vowels speech dataset with three different missingness levels and outperform the competitor in two out of three instances, failing to do so on the 50\% missing data setting.
The learned kernel splines that give the smallest distance to the data can be plotted to inspect which features contribute to the final prediction.

\section{Discussion}\label{sec:discussion}

In this paper we presented an alternative model for irregularly-sampled time series with missing data. Our method outperforms the competitors, such as the ODE-based models and transformers, on a variety of tasks. We also offer the ability to inspect the inner workings of our layers, which can be interpretable as they learn the shapes in the data.
One limitation of our work is the memory footprint which may be larger than in the other models depending on the number of learnable kernels. We did not, however, face any issues in our experiments because of this.

In the future, we can utilize more of the spline properties. For example, we could use regularized splines that do not pass through all the data points to drop the outliers. Further, we can have segments of varying sizes, acting on the time series simultaneously to obtain the hierarchical features, allowing learning at different scales.

\bibliographystyle{icml2022}
\bibliography{references.bib}

\end{document}